\documentclass[11pt]{article}

\usepackage[final]{acl}

\usepackage{times}
\usepackage{latexsym}
\usepackage[T1]{fontenc}
\usepackage[utf8]{inputenc}
\usepackage{microtype}
\usepackage{inconsolata}
\usepackage{graphicx}

\usepackage{amsmath}
\usepackage{amssymb}
\usepackage{booktabs}
\usepackage{multirow}
\usepackage{placeins}
\usepackage[most]{tcolorbox}

\title{Discriminative World Models for Web Agents}

\author{
\textbf{Kelvin Li}\textsuperscript{1}\thanks{Equal contribution.} \quad
\textbf{Dhruv Pendharkar}\textsuperscript{1}\footnotemark[1] \quad
\textbf{Anish Pahilajani}\textsuperscript{2} \quad
\textbf{Chuyi Shang}\textsuperscript{1} \quad
\textbf{Leon Oks}\textsuperscript{3} \\
\textbf{Leonid Karlinsky}\textsuperscript{4} \quad
\textbf{Rogerio Feris}\textsuperscript{2} \quad
\textbf{Trevor Darrell}\textsuperscript{1} \quad
\textbf{Roei Herzig}\textsuperscript{1} \\
\\[-0.5em]
\normalfont
\textsuperscript{1}University of California, Berkeley \quad
\textsuperscript{2}MIT-IBM Watson AI Lab \\
\textsuperscript{3}Cal Poly San Luis Obispo \quad
\textsuperscript{4}Xero
}

\begin{document}
\maketitle

\begin{abstract}
Recent web agents use world models for test-time action selection by sampling candidate actions, predicting the resulting web states, and ranking them with a ranker model or a Process Reward Model (PRM). These world models are typically trained via supervised next-state prediction to generate fixed representations like HTML or AXTree snapshots. However, this objective is misaligned with the downstream ranker, which relies on predicted states being discriminative across candidates to accurately score them. To address this, we introduce predicted-state matching, a training objective where the predicted representation must distinguish the true resulting state from those reached by alternative actions.  We train these models using a branching web-agent dataset derived from WebArena Go-Browse trajectories, where every decision point contains multiple alternative actions and their resulting states. Experiments on our held-out predicted-state matching benchmark show that our approach outperforms world models trained with supervised next-state prediction. We further show that our approach improves PRM-style action ranking on WebPRMBench compared with action-only PRMs and PRMs augmented with supervised-next-state world models. Finally, on WebArena-Lite, using our world model for test-time action selection improves end-to-end task success. Our project page is available at: https://dhruvpendharkar.github.io/dwm/.
\end{abstract}

\section{Introduction}
\label{sec:intro}

\begin{figure*}[t]
    \centering
    \includegraphics[width=0.88\textwidth]{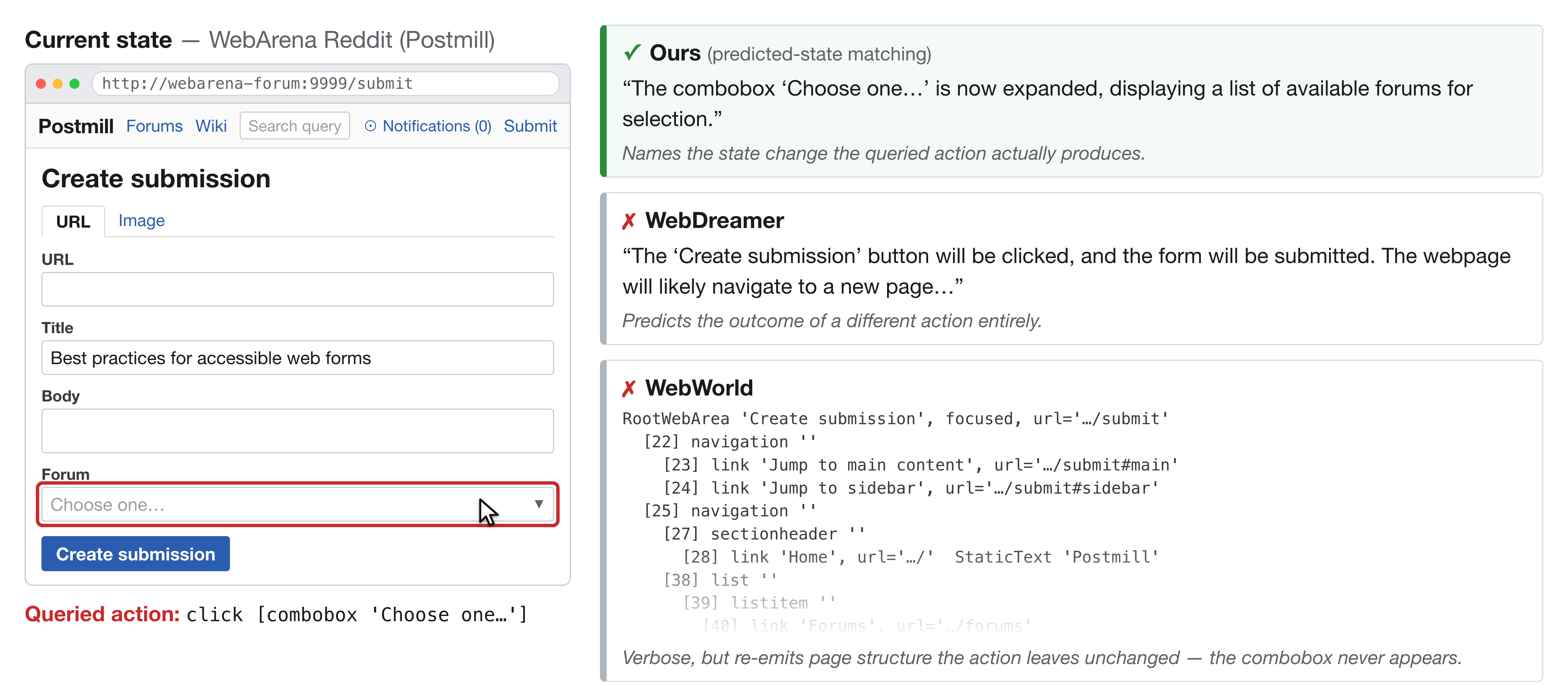}
    \caption{
    Qualitative example of predicted next-state representations at a single decision point in the WebArena Reddit environment. Given the current state and queried action, our predicted-state-matching world model identifies the specific state change produced by the action. A textual-summary method, WebDreamer~\citep{webdreamer}, instead describes the outcome of a different action, while a full structured-state method, WebWorld~\citep{webworld}, emits a long AXTree-style representation that largely repeats page structure unchanged by the queried action. The browser pane is rendered from the recorded accessibility tree.
    }
    \label{fig:qualitative_example}
\end{figure*}

Recent advances in large language models have made it possible to build agents that interact with the web through natural-language instructions and browser actions~\citep{seeact,webvoyager,ui-tars,molmoweb}. However, web navigation remains challenging because it is a partially observed, multi-step decision-making problem: an agent must choose actions based on the current page and interaction history, while reasoning about how each action will affect future steps. Most prior work relies on a reactive next-action prediction paradigm, where agents directly select an action based on the current observation and history~\citep{seeact,webvoyager,ui-tars,molmoweb}. Although straightforward, this approach does not explicitly compare alternative actions at a given state. To address this limitation, recent methods explore test-time action selection: the agent proposes multiple possible next actions, then uses a process reward model (PRM) or ranker to choose among them~\citep{webshepherd,webarbiter}. Model-based planning extends this paradigm by augmenting candidate actions with their predicted next states~\citep{webdreamer,wma,webevolver,webworld}. This allows the agent to explicitly evaluate and compare alternative paths through their expected consequences prior to execution.

This planning paradigm raises a basic question: what should the predicted next state represent? Current web world models typically rely on supervised next-state prediction, training the model to generate predefined formats like textual summaries, HTML, or AXTree snapshots~\citep{webdreamer,wma,webevolver,webworld}. This approach ties the training objective to arbitrary choices of state format. However, as illustrated in Figure~\ref{fig:qualitative_example}, a representation optimized to match a fixed next-state target may still miss the action-relevant differences that distinguish one action's consequence from another. For example, textual-summary methods such as WebDreamer are supervised to reproduce a particular compressed description of the next state, which may omit the specific change needed to distinguish the queried action from alternatives. In contrast, full structured-state methods such as WebWorld generate AXTree-style representations in which the relevant change can be buried among large amounts of unchanged page structure. When the world model is used for action selection, the predicted next state should help distinguish the outcomes of competing actions from the same state. We therefore argue that the relevant objective is not to recreate a specific target representation, but to produce predicted next-state representations that are discriminative among the alternative outcomes induced by candidate actions. 

\begin{figure*}[t]
    \centering
    \includegraphics[width=0.70\textwidth]{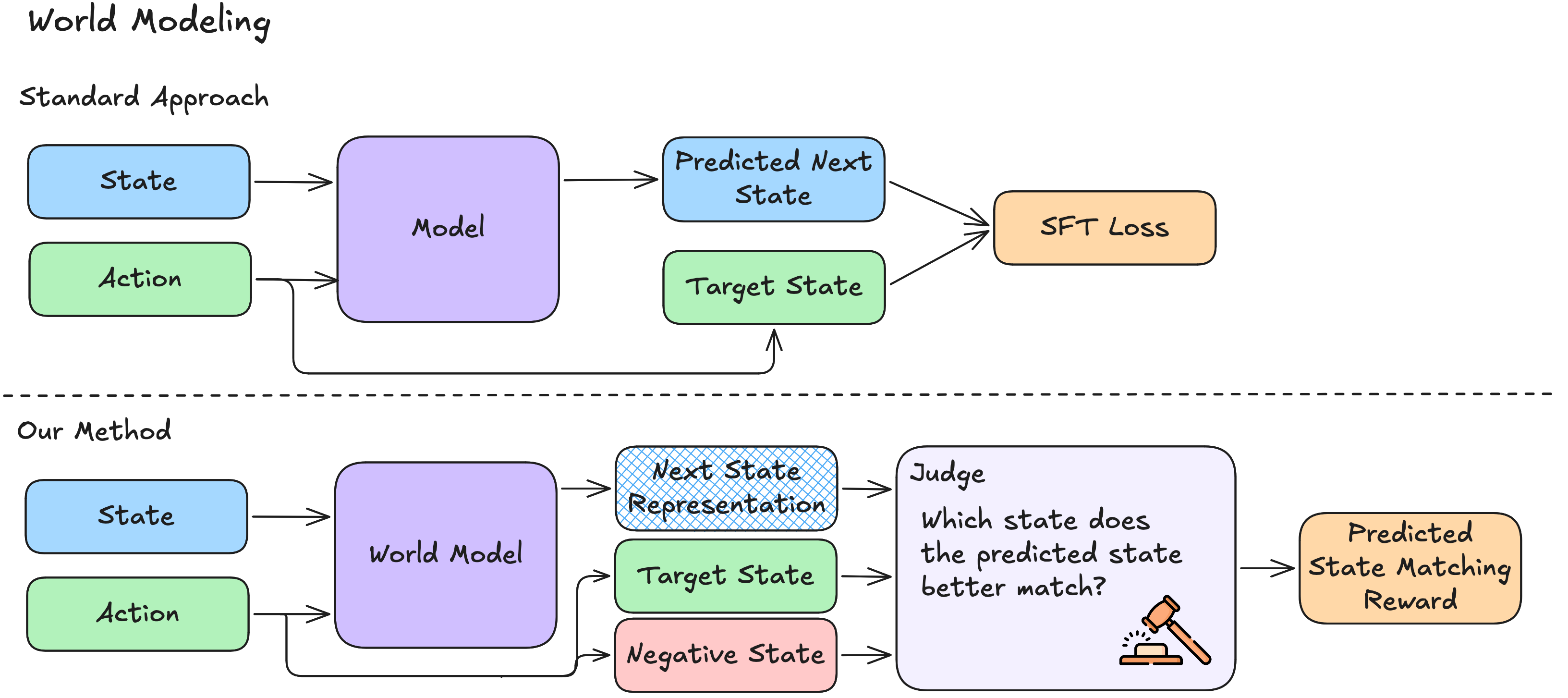}
    \caption{
    Supervised next-state prediction trains a world model to generate a fixed target representation. Predicted-state matching instead trains the generated representation to distinguish the true resulting state of a queried action from alternative next states reached by other actions.
    }
    \label{fig:world_modeling_overview}
\end{figure*}

We address this with predicted-state matching (Figure~\ref{fig:world_modeling_overview}). Given a current state and a candidate action, the model generates a next-state representation. Rather than forcing the model to generate a fixed target string, we train and evaluate this representation through a matching task, where a fixed judge uses the generated representation to distinguish the true resulting state of the queried action from alternative resulting states reached at the same decision point. This establishes a representation-agnostic objective for world modeling: a predicted representation is valuable specifically because it preserves the features that distinguish one action's consequence from another. Because the output is not tied to a rigid target format, the model can dynamically adapt its level of detail and abstraction across different tasks, pages, and actions.

To support this framework, we construct a branching web-agent dataset derived from Go-Browse trajectories in WebArena environments~\citep{gobrowse,webarena}. Unlike existing web-agent datasets that are organized as linear trajectories, our data contains local decision points with multiple possible actions from the same state and their resulting next states. This structure provides the supervision needed to train world models to distinguish the effects of different actions, rather than imitate a single gold path or generate a fixed state format. It also enables a held-out predicted-state matching benchmark for web world models: models are evaluated by whether their predicted next-state representation matches the correct resulting state among alternatives from the same decision point.

We evaluate our approach at three levels. First, on our held-out predicted-state-matching benchmark, our model achieves higher matching accuracy than existing web world models and a data-matched baseline trained with supervised next-state prediction. Second, we examine the downstream utility of these predictions for action ranking on WebPRMBench~\citep{webarbiter}. In model-based test-time action selection, predicted outcomes are useful only if they help a PRM or ranker choose among candidate actions before execution. WebPRMBench directly evaluates this ranking objective: holding the task context and candidate actions fixed, we compare action-only rankers with rankers augmented either with fixed-format next-state predictions or with representations produced by our predicted-state-matching world model. We evaluate both trained reward models and frozen rankers. Third, we evaluate end-to-end agent performance on WebArena-Lite~\citep{webarena-lite, webarena}, showing that incorporating our world model into test-time action selection improves task success.

We summarize our contributions as follows: (i) We formulate web world modeling as learning discriminative next-state representations that distinguish the outcomes of competing actions rather than generating fixed-format targets, introducing a representation-agnostic training objective called predicted-state matching. (ii) We introduce a branching web-agent dataset derived from Go-Browse trajectories, with local decision points containing alternative executable actions and their resulting states. This dataset supports training predicted-state-matching world models and provides a new representation-agnostic benchmark for evaluating whether a predicted next-state representation is distinguishable from the states reached by alternative actions from the same decision point. (iii) We show that our approach improves predicted-state matching over world models trained with supervised next-state prediction, that our predicted next-state representations improve PRM-style action ranking on WebPRMBench compared with action-only PRMs and supervised-next-state world model baselines, and that using our world model for test-time action selection improves end-to-end task success on WebArena-Lite.
\section{Related Work}
\label{sec:related}

\paragraph{Web agents.}
Web agents aim to complete natural-language tasks by interacting with websites through browser actions. Existing benchmarks study this problem across static offline examples~\citep{mind2web}, self-hosted web environments~\citep{webarena,visualwebarena,miniwob, miniwob++}, and live online websites~\citep{webvoyager,browsecomp,online_mind2web, mind2web-live}. These settings highlight the difficulty of web navigation as a partially observed, multi-step decision-making problem. Much work addresses this problem with direct next-action prediction, where the agent repeatedly selects one action given the task instruction, interaction history, and current observation until task completion~\citep{seeact,webvoyager,ui-tars,molmoweb}. While widely used, this reactive formulation does not explicitly compare plausible alternatives available at the same state. Recent work therefore explores test-time action selection, where an agent proposes multiple possible next actions and uses a ranker or PRM to select the best one~\citep{webshepherd,webarbiter}. Model-based planning extends this setup by using a world model to predict the next state that would result from each proposed action, allowing the ranker to compare predicted outcomes before selecting which action to execute~\citep{webdreamer,wma,webevolver,webworld}.

\paragraph{World models for web agents.}
Recent work in model-based planning uses world models for test-time action selection by predicting the next state resulting from proposed actions and using these predictions to rank and select an action~\citep{webdreamer,wma,webevolver,webworld}. These methods define the predicted state in fixed formats: natural-language state changes or transition abstractions~\citep{webdreamer,wma}, and structured web observations such as page states, HTML, or accessibility trees~\citep{webevolver,webworld}. These approaches show that predicted next states can improve action selection, but they train world models with supervised next-state prediction to generate a chosen state format. We argue that, for action selection, a predicted next state should help distinguish the outcomes of competing actions from the same current state, rather than only reproduce a chosen state format. Existing supervised next-state prediction objectives do not train or evaluate this property. We instead train and evaluate predicted next-state representations by whether they make the correct resulting state distinguishable from alternatives at the same decision point, and test whether these representations improve downstream action ranking.

\paragraph{Process reward models for web agents.}
PRMs provide step-level signals for test-time action selection in web navigation. Web-Shepherd trains a web PRM from step-level preference pairs and checklist annotations, while WebArbiter formulates WebPRM as structured reasoning with a preference verdict~\citep{webshepherd,webarbiter}. These methods improve action ranking, but they score candidate actions from the task context, interaction history, current observation, and action itself, without explicitly predicting the next state that each action would induce. Our work is complementary: we study whether PRMs can be improved by augmenting candidate actions with predicted next-state representations, and whether world models trained with predicted-state matching provide more useful representations than world models trained with supervised next-state prediction.

\paragraph{Web-agent data and evaluation.}
Most web-agent datasets are organized as linear trajectories, where a task is paired with a sequence of observations and actions~\citep{mind2web,weblinx,explorer,molmoweb}. This format supports both supervised fine-tuning of reactive web agents and supervised next-state prediction, but it records only the action taken along a single path. It therefore provides limited supervision for local decision making: at a given state, an agent must choose among several plausible actions, yet a linear trajectory does not show what would have happened under the alternatives. WebPRM datasets address part of this issue by providing scores or preference labels over candidate actions from the same state~\citep{webshepherd,webarbiter}. However, this supervision indicates which actions are preferred, but not the state changes that make one action preferable to another. For world models and world-aware ranking, this missing information is crucial, because comparing actions requires understanding how different actions will change the state differently. We build on Go-Browse~\citep{gobrowse} by organizing web-agent data around local decision points: for each state, we include multiple possible actions and the corresponding next states reached by those actions. This branching structure supports training and held-out evaluation through predicted-state matching, while downstream action-ranking performance is evaluated on WebPRMBench~\citep{webarbiter}.
\section{Method}
\label{sec:method}

We train web world models to produce predicted next-state representations that distinguish the outcomes of competing actions. As illustrated in Figure~\ref{fig:world_modeling_overview}, instead of supervising the model to generate a fixed target such as a textual transition summary or DOM/HTML/AXTree snapshot, we use \emph{predicted-state matching}: given a current state and a queried action, the predicted next-state representation is trained to be discriminative against the next states reached by alternative actions from the same decision point. We then evaluate these representations through predicted-state matching, PRM-style action ranking, and end-to-end agent execution.

\subsection{Problem setup and branching data}
\label{sec:method_setup}

We consider web navigation as a partially observed decision-making problem. At each step \(t\), the agent receives a task instruction \(I\), has an interaction history \(h_t\), observes the current web state \(s_t\), executes an action \(a_t\), and transitions to a next state \(s_{t+1}\).

Our method requires local decision points with multiple candidate actions from the same current state. We represent a decision point as
\[
    \mathcal{D}_t = \{(a_t^i, s_{t+1}^i)\}_{i=1}^{K},
\]
where \(a_t^i\) denotes the \(i\)-th candidate action and \(s_{t+1}^i\) is the next state reached after executing that action from \(s_t\). This branching structure exposes how different candidate actions from the same state lead to different next states.

\begin{figure}[t]
    \centering
    \includegraphics[width=\linewidth]{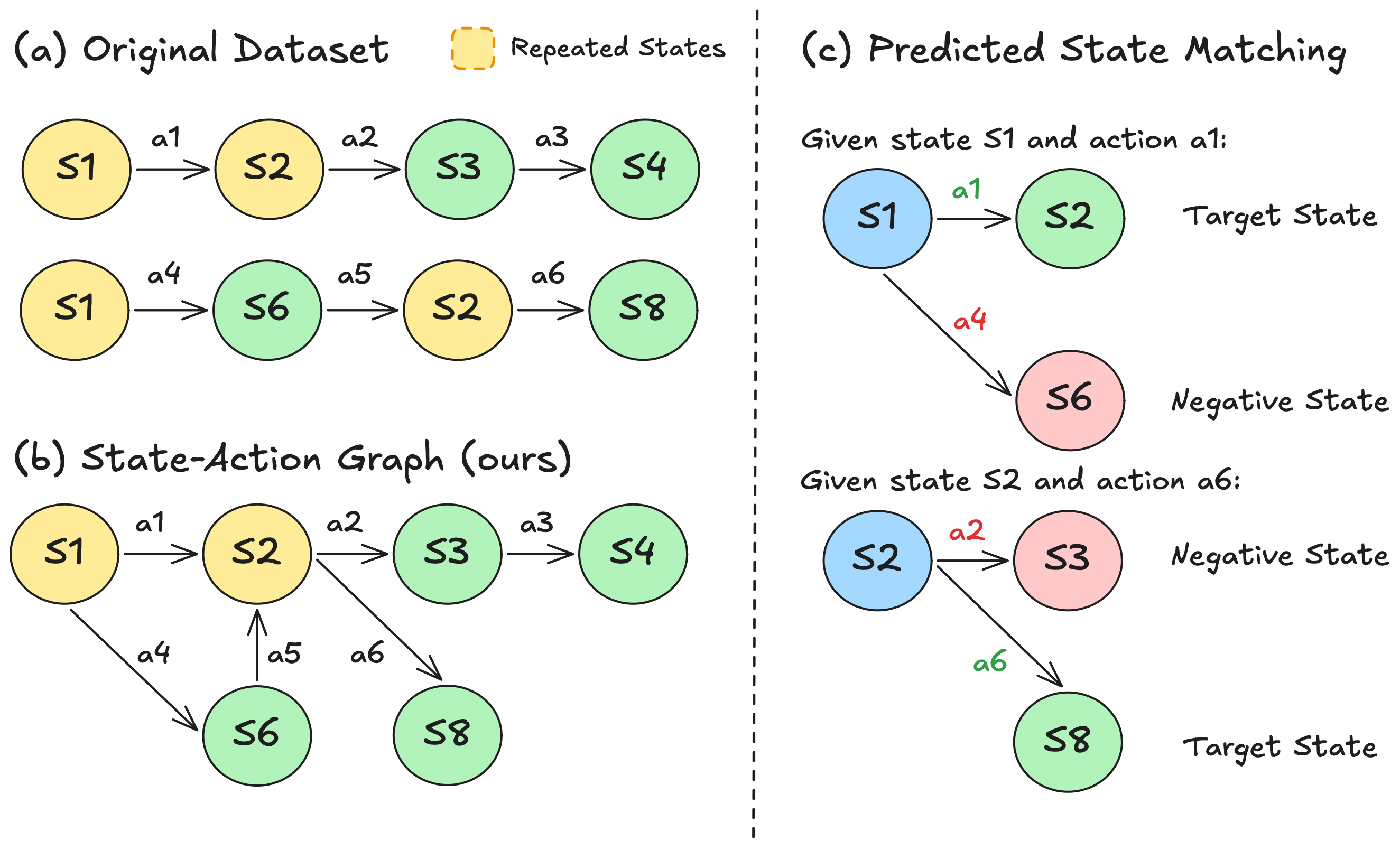}
    \caption{
    Construction of branching decision points from Go-Browse trajectories. 
    (a) The original data is organized as linear trajectories, but some web states are repeated across trajectories.
    (b) We merge repeated states into a state-action graph, where outgoing edges from the same state correspond to different executed actions and their observed next states.
    (c) For predicted-state matching, each branching state yields pairwise examples: given a queried action, the world model must produce a representation that distinguishes its true resulting state from an alternative next state reached by another action from the same current state.
    }
    \label{fig:branching_data}
\end{figure}

We construct such decision points from Go-Browse trajectories collected in WebArena environments~\citep{gobrowse,webarena}. As shown in Figure~\ref{fig:branching_data}(a), Go-Browse is organized as linear trajectories, but different trajectories can revisit the same browser states. We merge these repeated states into a state-action graph, where each node is a browser state represented by its accessibility tree and each directed edge is an executed action with its observed next state. This aggregation turns shared states with multiple outgoing actions into branching decision points, as shown in Figure~\ref{fig:branching_data}(b).

For predicted-state matching, we convert each branching decision point into pairwise examples, as shown in Figure~\ref{fig:branching_data}(c). Each example contains an instruction \(I\), history \(h_t\), current state \(s_t\), a queried action \(a_t^{\mathrm{qry}}\), its true resulting next state \(s_{t+1}^{\mathrm{qry}}\), and an alternative next state \(s_{t+1}^{\mathrm{alt}}\) reached by another action from the same decision point. The queried and alternative actions are not treated as good or bad actions; the objective is only to determine whether the predicted representation distinguishes the state caused by the queried action from the state caused by an alternative action.

The final world-modeling data contains 7,730 branching decision points from 2,839 trajectories, yielding 30,920 pairwise next-state matching examples. We split the data into training and evaluation sets with domain stratification to keep the WebArena domain distribution similar across splits.

\subsection{Predicted-state matching}
\label{sec:method_next_state}

Given an input \((I,h_t,s_t,a_t^{\mathrm{qry}})\), the world model generates a textual predicted next-state representation:
\[
    \hat{z}_{t+1}^{\mathrm{qry}}
    \sim
    \pi_\theta(\cdot \mid I,h_t,s_t,a_t^{\mathrm{qry}}).
\]
We use \(\hat{z}\) rather than \(\hat{s}\) to emphasize that the model output is a representation of the predicted next state, not the environment state. The representation is textual for interpretability and downstream compatibility with language-model-based ranking, but it is not supervised to match a predefined state string. 

To determine whether \(\hat{z}_{t+1}^{\mathrm{qry}}\) makes the correct next state distinguishable from alternatives, we use a matching judge \(J\). During training, we instantiate \(J\) with Qwen3-32B~\citep{qwen3}. The judge receives the predicted representation and two candidate next states:
\[
    \left(
    \hat{z}_{t+1}^{\mathrm{qry}},
    s_{t+1}^{\mathrm{qry}},
    s_{t+1}^{\mathrm{alt}}
    \right),
\]
with candidate order randomized. The judge must select which candidate state matches the predicted representation. Importantly, the judge does not receive the instruction, history, current state, or queried action. Therefore, the prediction is successful only if the generated representation itself contains enough information to identify the correct resulting state.

The matching reward is
\[
    R_{\mathrm{match}}
    =
    \mathbf{1}
    \left[
    J(\hat{z}_{t+1}^{\mathrm{qry}}, s_{t+1}^{\mathrm{qry}}, s_{t+1}^{\mathrm{alt}})
    =
    s_{t+1}^{\mathrm{qry}}
    \right].
\]
We also use a format reward \(R_{\mathrm{fmt}}\) to encourage well-formed textual outputs and discourage degenerate representations. We define \(R_{\mathrm{fmt}}=1\) if the model generates a non-empty state within valid \texttt{<predicted\_state>} XML tags, and \(R_{\mathrm{fmt}}=0\) otherwise. The total sequence-level reward is
\[
    R
    =
    R_{\mathrm{match}}
    +
    \lambda_{\mathrm{fmt}} R_{\mathrm{fmt}}.
\]
We set \(\lambda_{\mathrm{fmt}}=0.4\) in our main experiments. Additional implementation details and an ablation on \(\lambda_{\mathrm{fmt}}\) are provided in Appendix~\ref{app:training}.

This objective is representation-agnostic: the model is not trained to reproduce any fixed target representation of the next state. Ground-truth next states are used only as candidates for matching. The model is rewarded when its predicted representation distinguishes the true resulting state from alternatives induced by other actions at the same decision point.

\subsection{Using predicted next-state representations for action ranking}
\label{sec:method_ranking}

After training the world model with predicted-state matching, we use its generated representations as inputs to downstream PRMs. Given a candidate action \(a_t^i\), the world model generates
\[
    \hat{z}_{t+1}^i
    \sim
    \pi_\theta(\cdot \mid I,h_t,s_t,a_t^i).
\]
A downstream PRM or ranker then predicts a preference verdict from the current context and a pair of candidate actions, optionally augmented with predicted next-state representations:
\[
    \hat{y}
    \sim
    r_\phi\!\left(
    \cdot \mid
    I,h_t,s_t,
    (a_t^1,\hat{z}_{t+1}^1),
    (a_t^2,\hat{z}_{t+1}^2)
    \right),
\]
where \(\hat{y}\in\{1,2\}\) denotes the preferred candidate. In the no-state setting, the \(\hat{z}_{t+1}^i\) terms are omitted.

This interface lets us compare action-only ranking, ranking with fixed-format next-state predictions from supervised world models, and ranking with representations from our predicted-state-matching world model. In later experiments, we keep the preference-training and ranking setup fixed and vary the source of next-state information: action-only PRMs omit \(\hat{z}_{t+1}^i\), supervised next-state-prediction baselines use representations from world models trained to generate fixed target formats, and our method uses representations from the world model trained with predicted-state matching.
\section{Evaluation}
\label{sec:evaluation}

We evaluate our approach at three levels. First, we evaluate world-model quality on a held-out predicted-state matching benchmark, where each model generates a next-state representation for a queried action and a judge determines if it identifies the true resulting state among alternatives from the same decision point. We evaluate with judges from multiple model families to test whether the learned representations generalize beyond the judge used during training. Second, we evaluate whether these predicted representations improve downstream action ranking on WebPRMBench. WebPRMBench evaluates whether a PRM or ranker can identify the preferred action among alternatives from the same state; in our experiments, we augment candidate actions with predicted next-state representations and test whether this additional information improves ranking. We consider both trained and frozen reward-model settings. Third, we evaluate full agent trajectories on WebArena-Lite, testing whether using our world model for test-time action selection improves end-to-end task success.

\subsection{Predicted-state matching}
\label{sec:exp_state_matching}

We first evaluate world models in isolation using the predicted-state matching setup from Section~\ref{sec:method_next_state}. The evaluation uses the held-out split of our branching world-modeling data and covers five WebArena domains: Shopping, CMS, Reddit, GitLab, and Map. For each example, the evaluated model receives the task instruction, interaction history, current state, and a queried action, and generates a textual predicted next-state representation. For our primary evaluation, a fixed Qwen3-32B~\citep{qwen3} matching judge receives only this representation and two candidate next states: the true resulting state of the queried action and an alternative state reached by another action from the same decision point. The candidate order is randomized, and the judge must identify which candidate state matches the generated representation.

This benchmark is representation-agnostic: models are not evaluated by exact match to a predefined next-state string. Instead, a predicted next-state representation is correct if it distinguishes the state induced by the queried action from an alternative state induced by another action from the same current state. For the primary evaluation, every method receives the same task context and queried action and is evaluated with the same fixed Qwen3-32B matching judge; only the model used to generate the predicted next-state representation changes. We report two-way matching accuracy.

\paragraph{Baselines.}
We compare four groups of methods. First, we evaluate proprietary and open-source language models prompted to generate the predicted next-state representation. Second, we evaluate representative existing web world models trained with supervised next-state prediction, including WebDreamer-7B~\citep{webdreamer} and WebWorld-8B~\citep{webworld}. WebDreamer-7B represents world models that predict natural-language transition summaries, while WebWorld-8B represents recent structured-state world models. Third, we include a data-matched supervised next-state-prediction baseline trained on the same branching data, but supervised to generate fixed AXTree targets. This baseline controls for exposure to the same data while preserving the standard supervised next-state prediction objective. Fourth, we evaluate our Qwen3-8B world model trained with predicted-state matching. To test robustness to the choice of matching judge, we additionally re-evaluate our model, WebDreamer-7B, and WebWorld-8B using GPT-4o and Llama-3.1-70B-Instruct
, neither of which is used during training.

\begin{table*}[t]
\centering
\small
\renewcommand{\arraystretch}{0.90}

\caption{
Predicted-state matching accuracy on the held-out benchmark using Qwen3-32B as the matching judge.
Overall is micro-averaged.
}
\label{tab:state_matching}

\begin{tabular*}{0.90\textwidth}{
    @{\extracolsep{\fill}}
    @{}l
    cccccc
    @{}
}
\toprule
\textbf{Model}
& \textbf{Shopping}
& \textbf{CMS}
& \textbf{Reddit}
& \textbf{GitLab}
& \textbf{Map}
& \textbf{Overall} \\
\midrule

\multicolumn{7}{@{}l}{\textit{Proprietary LMs}} \\[-1pt]
GPT-4o
& 44.44 & 47.80 & 48.22 & 47.00 & 55.81 & 49.40 \\
GPT-4o-mini
& 48.15 & 46.70 & 47.72 & 47.00 & 53.49 & 48.80 \\

\midrule
\multicolumn{7}{@{}l}{\textit{Open-source LMs}} \\[-1pt]
Qwen3-4B
& 59.26 & 54.95 & 52.79 & 52.00 & 53.95 & 53.94 \\
Qwen3-8B
& 67.90 & 55.50 & 68.00 & 56.50 & 68.30 & 62.86 \\

\midrule
\multicolumn{7}{@{}l}{\textit{Existing world models}} \\[-1pt]
WebDreamer-7B
& 72.80 & 70.80 & 75.63 & 75.00 & 76.70 & 74.51 \\
WebWorld-8B
& \textbf{79.01} & 67.58 & 77.66 & 54.00 & 77.21 & 70.17 \\

\midrule
\multicolumn{7}{@{}l}{\textit{Data-matched SFT baseline}} \\[-1pt]
Full AXTree target
& 45.68 & 46.15 & 47.72 & 46.00 & 51.63 & 47.77 \\

\midrule
\multicolumn{7}{@{}l}{\textit{Ours}} \\[-1pt]
Predicted-state matching
& 77.78
& \textbf{77.47}
& \textbf{79.70}
& \textbf{82.50}
& \textbf{84.19}
& \textbf{80.80} \\

\bottomrule
\end{tabular*}

\end{table*}

\begin{table}[t]
\centering
\scriptsize
\setlength{\tabcolsep}{3.5pt}
\renewcommand{\arraystretch}{0.95}

\caption{
Predicted-state matching accuracy under different matching judges.
}
\label{tab:judge_robustness}

\begin{tabular}{@{}lccc@{}}
\toprule
\textbf{Model}
& \textbf{Qwen3-32B}
& \textbf{GPT-4o}
& \textbf{Llama-3.1-70B} \\
\midrule
WebDreamer-7B
& 74.51 & 76.91 & 76.10 \\
WebWorld-8B
& 70.17 & 74.06 & 76.00 \\
\midrule
Ours
& \textbf{80.80}
& \textbf{81.26}
& \textbf{79.31} \\
\bottomrule
\end{tabular}

\end{table}

\paragraph{Predicted-state matching results.}
Table~\ref{tab:state_matching} reports held-out predicted-state matching accuracy. Our model achieves the best overall accuracy, outperforming proprietary and open-source models, existing web world models, and the data-matched supervised next-state-prediction baseline. The data-matched supervised next-state-prediction baseline isolates the effect of the training objective: both models are fine-tuned from Qwen3-8B on the same data, but they use different training objectives. The baseline is supervised to generate a fixed AXTree target, whereas our model is trained through predicted-state matching. The improvement over this baseline indicates that matching-based supervision trains next-state representations that are better suited for distinguishing action-induced outcomes than supervised generation of a fixed state format. Additional analyses of predicted-state output length and training and data
efficiency are provided in Appendices~\ref{app:length}
and~\ref{app:training:efficiency}, respectively.

\paragraph{Robustness to the matching judge.}
Because Qwen3-32B is used to compute the matching reward during training, we test whether the learned representations remain discriminative under different judge models. As shown in Table~\ref{tab:judge_robustness}, our model outperforms both WebDreamer-7B and WebWorld-8B under Qwen3-32B, GPT-4o~\citep{gpt4o}, and Llama-3.1-70B-Instruct~\citep{grattafiori2024llama3herdmodels}. This indicates that the learned representations remain discriminative across judge model families rather than being specific to the judge used during training.

\subsection{WebPRMBench with trained reward models}
\label{sec:exp_webprm_trained}

We next evaluate whether next-state representations from our predicted-state-matching world model improve PRM-style action ranking in a trained-ranker setting. To test this, we train reward models on the WebPRM Collection~\citep{webshepherd} and evaluate them on WebPRMBench~\citep{webarbiter}. The WebPRM Collection provides preference-style training examples for web actions, while each WebPRMBench instance contains a task context, current web state, one preferred action, and four rejected alternatives. Following WebArbiter, we report Pairwise Accuracy, which measures whether the preferred action is ranked above a single rejected action, and Best-of-\(N\) (BoN) Accuracy, which requires the preferred action to be ranked above all rejected alternatives.

\paragraph{Baselines.}
Our core controlled comparison consists of the final three rows of Table~\ref{tab:webprm_trained}. All three use Qwen2.5-7B~\citep{qwen2.5} and the same answer-only preference-training setup as WebArbiter's full-data \emph{Instruct + SFT} baseline. Across these controlled settings, the reward-model backbone, training data, preference labels, and answer-only supervision objective are fixed; only the predicted next-state information provided to the ranker changes. Given the task context, interaction history, current state, and a pair of candidate actions, the reward model is trained to directly predict the preferred action, without intermediate reasoning supervision or reinforcement learning. For \emph{Direct (no state)}, we report WebArbiter's published \emph{Instruct + SFT} result. For \emph{+ WebWorld-8B states} and
\emph{+ state-matching states (ours)}, we append only a predicted next-state representation to the context for each candidate action, as described in Section~\ref{sec:method_ranking}, while keeping the reward-model training setup unchanged. \emph{+ WebWorld-8B states} uses predictions from frozen WebWorld-8B~\citep{webworld}, while \emph{+ state-matching states} uses predictions from our frozen predicted-state-matching world model. The world models are frozen during reward-model training. For broader context, Table~\ref{tab:webprm_trained} also includes LLM-as-judge and existing WebPRM results reported by WebArbiter; these rows are not part of the controlled comparison.

\begin{table*}[t]
\centering
\small
\caption{
WebPRMBench Pairwise and Best-of-\(N\) accuracy following WebArbiter.
The final three rows form our core controlled comparison: \emph{Direct (no state)}, \emph{+ WebWorld-8B states}, and \emph{+ state-matching states (ours)}.
All three rows use Qwen2.5-7B with the same answer-only preference-training setup and differ only in next-state information provided to the ranker.
\emph{Direct (no state)}, LLM-as-judge, and existing WebPRM results are reported by WebArbiter.
}
\label{tab:webprm_trained}
\resizebox{\textwidth}{!}{
\begin{tabular}{lcccccccccc}
\toprule
\multirow{2}{*}{\textbf{Model}}
& \multicolumn{2}{c}{\textbf{Mind2Web}}
& \multicolumn{2}{c}{\textbf{WebArena}}
& \multicolumn{2}{c}{\textbf{AssistantBench}}
& \multicolumn{2}{c}{\textbf{WorkArena}}
& \multicolumn{2}{c}{\textbf{Avg.}} \\
\cmidrule(lr){2-3}
\cmidrule(lr){4-5}
\cmidrule(lr){6-7}
\cmidrule(lr){8-9}
\cmidrule(lr){10-11}
& \textit{Pairwise} & \textit{BoN}
& \textit{Pairwise} & \textit{BoN}
& \textit{Pairwise} & \textit{BoN}
& \textit{Pairwise} & \textit{BoN}
& \textit{Pairwise} & \textit{BoN} \\
\midrule

\multicolumn{11}{l}{\textit{LLM-as-judge, proprietary language models}} \\
GPT-4o mini
& 81.74 & 50.92 & 78.23 & 56.72 & 89.17 & 73.33 & 81.43 & 46.70 & 82.64 & 56.92 \\
GPT-4o
& 79.99 & 52.62 & 84.58 & 66.67 & 85.83 & 66.67 & 84.33 & 55.19 & 83.68 & 60.29 \\
GPT-5
& 80.86 & 62.39 & 84.83 & 71.64 & 81.67 & 63.33 & 81.14 & 64.62 & 82.13 & 65.50 \\
Claude-3.7-Sonnet
& 80.20 & 57.90 & 82.80 & 64.10 & 81.50 & 61.30 & 82.10 & 60.60 & 81.65 & 60.98 \\

\midrule
\multicolumn{11}{l}{\textit{LLM-as-judge, open-source language models}} \\
Qwen2.5-3B
& 76.40 & 36.93 & 60.32 & 15.42 & 75.83 & 33.33 & 64.45 & 19.34 & 69.27 & 26.76 \\
Qwen2.5-7B
& 77.79 & 39.18 & 74.88 & 42.79 & 84.17 & 53.33 & 77.58 & 35.85 & 77.61 & 42.78 \\

\midrule
\multicolumn{11}{l}{\textit{Existing WebPRMs, 3B}} \\
WebShepherd-3B
& 87.50 & 65.21 & 68.16 & 41.29 & 66.67 & 46.67 & 50.00 & 21.23 & 68.08 & 43.60 \\
WebArbiter-3B
& 93.32 & 78.42 & 81.97 & 56.22 & 78.33 & 46.67 & 81.01 & 54.81 & 83.65 & 59.06 \\

\midrule
\multicolumn{11}{l}{\textit{Existing WebPRMs, 7B+}} \\
WebShepherd-8B
& 86.66 & 73.69 & 68.33 & 43.88 & 55.92 & 30.00 & 54.56 & 25.53 & 64.34 & 43.28 \\
WebArbiter-7B
& \textbf{97.07} & \textbf{89.53} & 88.43 & 68.66 & \textbf{89.17} & \textbf{70.00} & 82.09 & \textbf{70.19} & 89.19 & \textbf{74.60} \\

\midrule
\multicolumn{11}{l}{\textit{Controlled Qwen2.5-7B reward models}} \\
Direct (no state)
& 85.14 & 60.91 & 80.85 & 52.73 & 82.50 & 56.67 & 79.57 & 52.88 & 82.02 & 55.80 \\
+ WebWorld-8B states
& 89.74 & 73.20 & 84.32 & 65.60 & 86.66 & 67.41 & 81.21 & 64.32 & 85.48 & 67.63 \\
\textbf{+ state-matching states (ours)}
& 96.71 & 89.39 & \textbf{88.56} & \textbf{69.15} & 88.33 & \textbf{70.00} & \textbf{83.84} & 62.26 & \textbf{89.36} & 72.70 \\
\bottomrule
\end{tabular}
}
\end{table*}

\paragraph{Action-ranking results.}
Table~\ref{tab:webprm_trained} reports WebPRMBench action-ranking accuracy with trained reward models. In the controlled comparison (final three rows), augmenting the reward model with our predicted next-state representations improves over the \emph{Direct (no state)} baseline, showing that predicted action outcomes provide useful information beyond the candidate actions themselves. Our predicted states also outperform WebWorld-8B states under the same reward-model backbone and SFT training setup, indicating that representations learned through predicted-state matching are more useful for downstream action ranking than supervised next-state predictions. More broadly, our state-augmented reward model achieves competitive performance among the methods in Table~\ref{tab:webprm_trained}, outperforming the reported proprietary LLM-as-judge baselines as well as WebShepherd while being competitive with WebArbiter. We note that our training setup uses answer-only supervision, without additional structured reasoning supervision or reinforcement learning. In comparison, WebShepherd uses checklist-guided reward modeling, while WebArbiter uses principle-guided reasoning trained through reasoning distillation and reinforcement learning. These results suggest that explicitly providing predicted action outcomes can be a meaningful source of information for action ranking, even without adding a more elaborate reasoning or reinforcement-learning procedure.

\subsection{WebPRMBench with frozen reward models}
\label{sec:exp_webprm_frozen}

To separate the usefulness of predicted next-state representations from reward-model training, we repeat the same controlled comparisons as Section~\ref{sec:exp_webprm_trained} with frozen rankers. In this training-free setting, a frozen Qwen2.5 model receives the task context, current state, candidate actions, and optionally predicted next-state representations, and outputs which action is preferred.

\paragraph{Baselines.}
We evaluate Qwen2.5-3B and Qwen2.5-7B. For each model size, we compare three input settings: no predicted state, predicted states from WebWorld-8B, and predicted states from our predicted-state-matching world model. The no-state rows are reported by WebArbiter. For the state-augmented settings, the Qwen2.5 ranker remains frozen, and predicted next states are added to the input context; the two settings differ only in which frozen world model provides these states.

\begin{table*}[t]
\centering
\small
\caption{
Frozen-ranker evaluation on WebPRMBench.
A frozen Qwen2.5 model predicts the preferred action either directly with no predicted next-state information, or after augmentation with predicted states from a frozen world model.
The no-state results are reported by WebArbiter; the state-augmented rows are our evaluations using WebWorld-8B or our predicted-state-matching world model.
}
\label{tab:webprm_frozen}
\resizebox{\textwidth}{!}{
\begin{tabular}{lcccccccccc}
\toprule
\multirow{2}{*}{\textbf{Model}}
& \multicolumn{2}{c}{\textbf{Mind2Web}}
& \multicolumn{2}{c}{\textbf{WebArena}}
& \multicolumn{2}{c}{\textbf{AssistantBench}}
& \multicolumn{2}{c}{\textbf{WorkArena}}
& \multicolumn{2}{c}{\textbf{Avg.}} \\
\cmidrule(lr){2-3}
\cmidrule(lr){4-5}
\cmidrule(lr){6-7}
\cmidrule(lr){8-9}
\cmidrule(lr){10-11}
& \textit{Pairwise} & \textit{BoN}
& \textit{Pairwise} & \textit{BoN}
& \textit{Pairwise} & \textit{BoN}
& \textit{Pairwise} & \textit{BoN}
& \textit{Pairwise} & \textit{BoN} \\
\midrule

\multicolumn{11}{l}{\textit{Frozen Qwen2.5-3B ranker}} \\
No predicted state
& 76.40 & 36.93 & 60.32 & 15.42 & 75.83 & 33.33 & 64.45 & 19.34 & 69.27 & 26.76 \\
+ WebWorld-8B states
& 65.84 & 32.81 & 69.90 & 36.82 & 82.50 & 56.67 & 55.31 & 20.75 & 68.39 & 36.76 \\
\textbf{+ state-matching states (ours)}
& \textbf{76.77} & \textbf{42.01} & \textbf{71.02} & \textbf{36.82} & \textbf{76.67} & \textbf{56.67} & \textbf{71.82} & \textbf{36.32} & \textbf{74.07} & \textbf{42.96} \\

\midrule
\multicolumn{11}{l}{\textit{Frozen Qwen2.5-7B ranker}} \\
No predicted state
& 77.79 & 39.18 & 74.88 & 42.79 & 84.17 & 53.33 & 77.58 & 35.85 & 77.61 & 42.78 \\
+ WebWorld-8B states
& 73.09 & 44.69 & \textbf{77.74} & \textbf{55.22} & 82.50 & 60.00 & 69.10 & 46.23 & 75.61 & 51.53 \\
\textbf{+ state-matching states (ours)}
& \textbf{80.06} & \textbf{48.80} & 77.00 & 51.24 & \textbf{84.17} & \textbf{66.67} & \textbf{79.01} & \textbf{51.89} & \textbf{80.06} & \textbf{54.65} \\

\bottomrule
\end{tabular}
}
\end{table*}

\paragraph{Action-ranking results.}
Table~\ref{tab:webprm_frozen} reports WebPRMBench action-ranking accuracy with frozen rankers. Across both Qwen2.5-3B and Qwen2.5-7B, augmenting the frozen ranker with predicted next states improves over the no-state setting, showing that predicted action outcomes provide useful information even without reward-model training. Our predicted states also consistently outperform WebWorld-8B states, indicating that representations learned through predicted-state matching are more useful for downstream action ranking than supervised next-state predictions.

\subsection{End-to-end agent evaluation}
\label{sec:exp_end_to_end}

Finally, we evaluate whether our predicted next-state representations improve end-to-end agent performance on WebArena-Lite~\citep{webarena-lite,webarena}.

\paragraph{Baselines.}
We use GPT-4o~\citep{gpt4o} as the policy model and compare three inference-time settings. \emph{ReAct-style}~\citep{react} follows the standard web-agent loop: at each step, GPT-4o reasons over the current task context and interaction history, selects a single action, and executes it. \emph{Best-of-5 (Bo5)} instead has GPT-4o propose five candidate actions and then rank them to select one for execution, with GPT-4o serving as both the proposer and ranker. \emph{Bo5 + state matching} follows a similar setup: GPT-4o proposes five candidate actions, our frozen world model predicts a next-state representation for each candidate, and GPT-4o then selects which action to execute using both the candidate actions and their predicted next states. Because end-to-end web-agent results can vary with the implementation harness, we implement and evaluate all three settings within the same framework to enable a controlled comparison.

\paragraph{End-to-end results.}
\emph{ReAct-style} achieves \(13.94\%\) task success, closely matching the \(13.9\%\) GPT-4o result reported by WebRL~\citep{webrl}. \emph{Bo5} improves task success to \(21.82\%\), showing that sampling and ranking multiple candidate actions improves over single-action selection. \emph{Bo5 + state matching} further improves task success to \(28.48\%\), showing that predicted next-state information helps GPT-4o rank candidate actions more effectively.
\section{Conclusion}
\label{sec:conclusion}
We introduced predicted-state matching, a representation-agnostic objective for training web world models to produce next-state representations that distinguish competing action outcomes, together with a branching web-agent dataset for training and evaluation. Instead of supervising a fixed state format, predicted-state matching trains representations to distinguish the true resulting next state from alternatives reached by other actions at the decision point. Our model improves held-out matching accuracy over supervised next-state-prediction baselines, and its representations improve WebPRMBench action ranking with trained and frozen rankers. They also improve end-to-end task success on WebArena-Lite with model-based Best-of-5 action selection. Together, these results suggest that web world models should preserve action-relevant differences for downstream decision making rather than only reproduce a predefined next-state representation.
\FloatBarrier
\section{Limitations}
\label{sec:limitations}

Our predicted-state matching benchmark is constructed from Go-Browse trajectories in WebArena environments, and our end-to-end evaluation is conducted on WebArena-Lite using GPT-4o as the policy model. As a result, the generality of our findings across other web environments and policy models remains to be established. Evaluating the approach on more diverse benchmarks and models is an important direction for future work.

Predicted-state matching relies on language-model judges to determine whether a generated representation corresponds to the correct resulting state. We evaluate with Qwen3-32B, GPT-4o, and Llama-3.1-70B-Instruct and observe consistent improvements across judges, but matching accuracy remains a model-based proxy and may reflect limitations or biases shared by these judge models.

Our branching data is constructed from alternative actions and outcomes observed in Go-Browse trajectories. This provides executable action-outcome branches from shared web states, but does not enumerate every possible action that could have been taken at each decision point. Extending the framework to more densely explored branching data and a broader range of web environments could provide additional supervision for predicted-state matching.
\section{Ethical Considerations}
\label{sec:ethics}

We use and cite existing research artifacts, including Go-Browse, WebArena, WebPRM Collection, WebPRMBench, WebDreamer, WebWorld, and Qwen model families. Our use of these artifacts is for research evaluation and model development. Any released derived data, code, or model checkpoints should follow the licenses and terms of the original artifacts; when redistribution is not permitted, we will release scripts or processing instructions rather than the restricted artifacts themselves.

The data used in this work comes from web-agent benchmark environments and previously released trajectories. We do not collect new human-subject data or recruit human annotators. The benchmark states and trajectories are intended for research use, and our derived branching data is intended for research on web-agent world modeling and action ranking, not for deployment in unrestricted autonomous web agents.

We do not intentionally include personally identifying information. Since web states can contain natural-language content, we recommend that any released derived dataset be checked for personally identifying or sensitive content before distribution. Our experiments are limited to benchmark environments and do not execute actions on real user accounts or live personal data.

Our method is intended to improve world modeling and action selection for web agents, which could also increase the capabilities of autonomous web agents if deployed outside benchmark environments. Such systems may take unintended or harmful actions, particularly when interacting with real websites, user accounts, or sensitive information. Our experiments are restricted to benchmark environments and do not evaluate or advocate unrestricted autonomous deployment.

Throughout this project we utilized AI models for the purpose of writing code, specifically Claude Code and ChatGPT. All model outputs were verified and proofread by the authors to ensure they followed intended behavior.

\FloatBarrier
\bibliography{custom}

\clearpage
\appendix

\newtcolorbox{examplecontext}{
  breakable,
  colback=black!3, colframe=black!40, boxrule=0.4pt, arc=2pt,
  left=4pt, right=4pt, top=3pt, bottom=3pt,
  title={Context}, fonttitle=\bfseries\small, fontupper=\small,
}
\newtcblisting{statebox}[1]{
  listing only, breakable,
  colback=white, colframe=black!55, boxrule=0.4pt, arc=2pt,
  left=4pt, right=4pt, top=3pt, bottom=3pt,
  title={#1}, fonttitle=\bfseries\small,
  listing options={
    basicstyle=\ttfamily\footnotesize,
    breaklines=true, breakatwhitespace=false,
    columns=fullflexible, keepspaces=true, upquote=true,
  },
}

\newtcblisting{promptbox}[1][Prompt]{
  listing only, breakable,
  colback=black!3, colframe=black!40, boxrule=0.4pt, arc=2pt,
  left=4pt, right=4pt, top=3pt, bottom=3pt,
  title={#1}, fonttitle=\bfseries\small,
  listing options={
    basicstyle=\ttfamily\footnotesize,
    breaklines=true, breakatwhitespace=false,
    columns=fullflexible, keepspaces=true, upquote=true,
  },
}

\twocolumn[
  \begin{center}
    {\LARGE\bfseries Discriminative World Models for Web Agents\par}
    \vspace{0.6em}
    {\Large Supplementary Material\par}
    \vspace{1.2em}
  \end{center}
]

\noindent Here, we provide additional qualitative examples of predicted next-state representations (Appendix~\ref{app:qualitative}), an analysis of the length of the generated representations (Appendix~\ref{app:length}), training and implementation details for the world model and downstream reward models (Appendix~\ref{app:training}), and the full prompts used throughout our experiments (Appendix~\ref{app:prompts}).

\section{Qualitative Examples of Predicted States}
\label{app:qualitative}

This section provides qualitative examples of our predicted states as compared to WebDreamer and WebWorld.

\subsection{Example 1}

\begin{examplecontext}
 In this example, the current page is a GitLab To-Do List and the candidate action clicks the Files link for the A11Y Project repository.
\end{examplecontext}

\begin{statebox}{Ours}
"The page state will show the content of the 'Files' section of the A11Y Project repository, with the focus on the file listing or directory structure, as the 'Files' link was clicked. The URL will likely update to include a path to a specific file or directory, and the accessibility tree will reflect the new content, such as file names, folders, and navigation elements related to the selected file or directory."
\end{statebox}

\begin{statebox}{WebDreamer}
"The webpage will display a filter option that allows you to filter the to-do list to only show merge request related items."
\end{statebox}

\begin{statebox}{WebWorld (truncated)}
"\n\nRootWebArea 'To-Do List \u00b7 Dashboard \u00b7 GitLab', focused, url='http://ec2-3-148-123-246.us-east-2.compute.amazonaws.com:8023/dashboard/todos'
\n\t[53] banner ''
\n\t\t[54] link 'Skip to content', url='http://ec2-3-148-123-246.us-east-2.compute.amazonaws.com:8023/dashboard/todos#content-body'\n\t\tStaticText 'GitLab'
\n\t\t[60] link 'Dashboard', url='http://ec2-3-148-123-246.us-east-2.compute.amazonaws.com:8023/'
\n\t\t\t[61] image ''
\n\t\t[64] list ''
\n\t\t\t[65] listitem ''
\n\t\t\t\t[66] button '', hasPopup='menu', expanded=False
\n\t\t[107] list ''
\n\t\t\t[108] listitem ''
\n\t\t\t\t[112] image ''
\n\t\t\t\t[113] textbox 'Search GitLab'
\n\t\t\t\tStaticText '/'
\n\t\t[122] list ''
\n\t\t\t[123] listitem ''"

\end{statebox}

\subsection{Example 2}

\begin{examplecontext}
In this example the models are given an action of selecting which forum to make a post within the Reddit domain.
\end{examplecontext}

\begin{statebox}{Ours}
"The combobox 'Choose one...' is now expanded, displaying a list of available forums for selection."
\end{statebox}

\begin{statebox}{WebDreamer}
"The "Create submission" button will be clicked, and the form will be submitted. The webpage will likely navigate to a new page or display a confirmation message indicating that the submission has been successfully created."
\end{statebox}

\begin{statebox}{WebWorld (truncated)}
"RootWebArea 'Create submission', focused, url='http://ec2-3-148-123-246.us-east-2.compute.amazonaws.com:9999/submit'
        [22] navigation ''
                [23] link 'Jump to main content', url='http://ec2-3-148-123-246.us-east-2.compute.amazonaws.com:9999/submit#main'
                [24] link 'Jump to sidebar', url='http://ec2-3-148-123-246.us-east-2.compute.amazonaws.com:9999/submit#sidebar'
        [25] navigation ''
                [27] sectionheader ''
                        [28] link 'Home', url='http://ec2-3-148-123-246.us-east-2.compute.amazonaws.com:9999/'
                                StaticText 'Postmill'
                [38] list ''
                        [39] listitem ''
                                [40] link 'Forums', url='http://ec2-3-148-123-246.us-east-2.compute.amazonaws.com:9999/forums'
                        [41] listitem ''
                                [42] link 'Wiki', url='http://ec2-3-148-123-246.us-east-2.compute.amazonaws.com:9999/wiki'
                [52] searchbox 'Search query'
                [53] list ''
                        [54] listitem ''
                                [55] link 'Notifications (0)', url='http://ec2-3-148-123-246.us-east-2.compute.amazonaws.com:9999/notifications'
                                        [58] image ''
                        [59] listitem ''
                                [60] link 'Submit', url='http://ec2-3-148-123-246.us-east-2.compute.amazonaws.com:9999/submi"

\end{statebox}

\subsection{Example 3}

\begin{examplecontext}
In this example the world models are given a current page state in the ecommerce environment with the starting point of a home page and the action performed being clicking the catalog button.
\end{examplecontext}

\begin{statebox}{Ours}
"The page state will change to display the CATALOG section, with the focus on the CATALOG menu item. The content area will likely show product-related information or settings specific to the CATALOG section, such as product listings, categories, or inventory management options."
\end{statebox}

\begin{statebox}{WebDreamer}
"The product list will be displayed with more columns, providing a more detailed view of the products."
\end{statebox}

\begin{statebox}{WebWorld (truncated)}
"RootWebArea 'Catalog / Magento Admin', focused, url='http://ec2-3-148-123-246.us-east-2.compute.amazonaws.com:7780/admin/admin/catalog/'
        [148] link 'Magento Admin Panel', url='http://ec2-3-148-123-246.us-east-2.compute.amazonaws.com:7780/admin/admin/'
                [149] image 'Magento Admin Panel', url='http://ec2-3-148-123-246.us-east-2.compute.amazonaws.com:7780/static/version1742719628/adminhtml/Magento/backend/en_US/images/magento-icon.svg'
        [150] navigation ''
                [151] menubar '', orientation='horizontal'
                        [153] link '\ue604 DASHBOARD', url='http://ec2-3-148-123-246.us-east-2.compute.amazonaws.com:7780/admin/admin/dashboard/'
                                StaticText '\ue604'
                                StaticText 'DASHBOARD'
                        [156] link '\ue60b SALES', url='http://ec2-3-148-123-246.us-east-2.compute.amazonaws.com:7780/admin/admin/dashboard/#'
                                StaticText '\ue60b'
                                StaticText 'SALES'
                        [189] link '\ue608 CATALOG', url='http://ec2-3-148-123-246.us-east-2.compute.amazonaws.com:7780/admin/admin/catalog/'
                                StaticText '\ue608'
                                StaticText 'CATALOG'
                        [207] link '\ue603 CUSTOMERS', url='http://ec2-3-148-123-246.us-east-2.compute.amazonaws.com:7780/admin/admin/dashboard/#'
                                StaticText '\ue603'
                                StaticText 'CUSTOMERS'
                        [226] link '\ue609 MARKETING', url='http://ec2-3-148-123-246.us-east-2.compute.amazonaws.com:7780/admin/admin/dashboard/#'
                                StaticText '\ue609'
                                StaticText 'MARKETING'
                        [293] link '\ue602 CONTENT', url='http://ec2-3-148-123-246.us-east-2.compute.amazonaws.com:7780/admin/admin/dashboard/#'"

\end{statebox}

\clearpage

\section{Output Length Analysis}
\label{app:length}
Our model produces substantially shorter predicted-state representations than WebWorld-8B while achieving stronger performance on both predicted-state matching and downstream action ranking. Our predictions average \(91.6\) tokens per state, compared with \(412.7\) tokens for WebWorld-8B, excluding its additional reasoning tokens. As shown in Table~\ref{tab:state_matching}, our model achieves \(80.80\%\) predicted-state matching accuracy compared with \(70.17\%\) for WebWorld-8B. When used to augment the controlled Qwen2.5-7B reward model on WebPRMBench, our predicted states achieve \(72.70\%\) average Best-of-\(N\) accuracy compared with \(67.63\%\) for WebWorld-8B (Table~\ref{tab:webprm_trained}). Our model therefore produces substantially shorter representations than WebWorld-8B, which predicts full AXTrees, while achieving stronger performance on both benchmarks. This suggests that the gains arise from more effective state representations rather than greater output length.

\section{Training Details}
\label{app:training}

This section provides additional implementation and training details for the
predicted-state-matching world model described in Section~\ref{sec:method_next_state}.

\subsection{Implementation Details}
\label{app:training:details}
We optimize the world model using Group Relative Policy Optimization (GRPO)~\citep{grpo}, implemented with TRL. For each input, we sample 8 completions and compute group-relative advantages from their total sequence-level rewards. The resulting completion-level advantage is applied across its generated tokens through the standard GRPO policy objective. We use KL regularization with coefficient \(0.04\). During GRPO training, completions are sampled with a temperature of \(1.0\), while at inference time the world model uses greedy decoding; both use a maximum generation length of \(2{,}048\) tokens.

\subsection{Training and Data Efficiency}
\label{app:training:efficiency}
Our world model
is trained for 4,830 optimization steps with an effective batch size of 32, requiring 48.75 hours on eight
A100 GPUs—six for optimization and two for gen-
eration during GRPO training—for a total of 390
GPU-hours. In comparison, WebWorld-8B reports
1,568 A100 GPU-hours, while WebDreamer-7B
uses 64 H100 GPUs but does not report total GPU-
hours. Our model is also trained on only 30,920
pairwise examples constructed from 7,730 branch-
ing decision points, compared with over 3.1M
synthesized web interactions for WebDreamer-7B
and 1.06M trajectories for WebWorld-8B. Despite
this difference in scale, predicted-state matching
achieves stronger held-out matching accuracy, sug-
gesting that aligning the training objective with
action-outcome discrimination can be more impor-
tant than simply scaling supervised next-state pre-
diction.

\subsection{Format Reward Ablation}
\label{app:training:format_ablation}
We provide ablations on the value of \(\lambda_{\mathrm{fmt}}\) which is used to scale the format reward given in our RL training. All experiments were run with identical training steps and all other training details are identical to those described in Appendix~\ref{app:training:details}. The results are shown in Table~\ref{tab:format_reward_ablation}. We find that the predicted-state matching accuracy is relatively robust to the choice of \(\lambda_{\mathrm{fmt}}\).

\begin{table}[h]
\centering
\small
\renewcommand{\arraystretch}{0.95}

\caption{
Effect of the format-reward weight \(\lambda_{\mathrm{fmt}}\) on predicted-state matching accuracy.
}
\label{tab:format_reward_ablation}

\begin{tabular}{@{}lr@{}}
\toprule
\textbf{Value of \(\lambda_{\mathrm{fmt}}\)}
& \textbf{Predicted-state matching accuracy} \\
\midrule
0.2 & 79.0\% \\
0.4 & 80.8\% \\
0.6 & 78.9\% \\
1.0 & 79.2\% \\
\bottomrule
\end{tabular}

\end{table}

\section{Prompts}
\label{app:prompts}

\subsection{World Model Prompt}

\begin{promptbox}
"You are a web browser world model. "
"Given the current page state (as an accessibility tree), prior action history "
"(when provided), and an action that was performed, predict what the next page "
"state would look like. Use the history to understand the broader context of "
"the task, but focus on the effect of the most recent action. "
"Do not hedge - commit to a specific prediction.\n\n"
"Respond in this exact format:\n"
"<predicted_state>Your prediction of the next page state</predicted_state>"
\end{promptbox}

\subsection{Matching Judge Prompt}

\begin{promptbox}
"You are evaluating web browser state predictions. "
"You will be shown a predicted next page state and two candidate actual next "
"states (as accessibility trees). Determine which candidate better matches "
"the predicted state.\n\n"
"Respond in this exact format:\n"
"<answer>A or B</answer>"
\end{promptbox}

\end{document}